# GENETIC ALGORITHMS AND ITS USE WITH BACK-PROPAGATION NETWORK

*Eng. Ayman M Bahaa Eldeen Sadeq, Prof. Dr. Abdel-Moneim A. Wahdan,*
*Prof. Dr. Hani M. K. Mahdi*
*Faculty of Engineering, Ain Shams University*

**Abstract:**
Genetic algorithms are considered as one of the most efficient search techniques. Although they do not offer an optimal solution, their ability to reach a suitable solution in considerably short time gives them their respectable role in many AI techniques. This work introduces genetic algorithms and describes their characteristics. Then a novel method using genetic algorithm in best training set generation and selection for a back-propagation network is proposed. This work also offers a new extension to the original genetic algorithms

ملخص

تعتبر الخوارزمات الجينية واحدة من أفضل طرق البحث من ناحية الأداء . فبالرغم من أن استخدام هذه الطريقة لا يعطي الحل الأمثل، إلا أن وصول الخوارزم لحل مناسب في زمن قصير جداً يعطي هذه الطريقة دورها الهام في كثير من عمليات الذكاء الاصطناعي. وفي هذا البحث نقوم بعرض سريع للخوارزمات الجينية ثم نقوم بعد ذلك بتقديم طريقة مبتكرة لاستخدام الخوارزمات الجينية لانتقاء أفضل مجموعة تعليم لشبكة عصبية من نوع الانتشار العكسي . وللوصول لهذه الطريقة تم اقتراح تطوير للخوارزمات الجينية الأصلية

## 1. Introduction

This research targets the problem of finding the best training set for a back-propagation network. The choice of this training set highly affects both the training time and the efficiency of the network at recall time. The chosen set must be a super set for most, if not all, target domain of patterns. That means, when the network is trained by this set, it can recognize most of, if not all, the target domain patterns.

A solution to the problem is to select a huge set of patterns to train the network with. In this case the coverage of the training process is guaranteed to some extent.

The problem for this solution is the extremely large training time needed and the divergence problem. For large data sets, the network may suffer contradicting features that results in a divergence or a local minimum stuck, while these features may be of unnecessary role in the overall recognition process. The search of a set that holds all the *important* features of all the target domain patterns is the perfect solution for this problem. The importance of features, which are automatically extracted by the network, can be judged by the final recognition rate of a relatively large test set. The testing process takes a very few amount of time since each pattern is propagated once in only one direction unlike the training process which takes very large number of iterations in both forward and backward directions to accomplish weights adjustment.

The ability of genetic algorithms of finding solutions to optimization problems in a very fast way is considered a candidate solution to the training set selection problem. Starting with a set of training sets, and then applying the genetic algorithms to mate and mutate sets to generate new sets, which were never existing, can lead to the desired training set. This ability is studied in this paper and an extension to the genetic algorithms is proposed. A suitable fitness criterion is also proposed.

## 2. Genetic algorithms

In this section a brief description of genetic algorithm and its characteristics is introduced [1]. Holland (1969, 1975) proposed the use of genetic algorithm as an efficient search mechanism in artificially adaptive systems. This procedure was designed to incorporate specific operators, emphasizing especially crossover and inversion. Crossover was defined (Holland, 1975) as taking two coding sequences and exchanging the set of attributes that follow a randomly chosen position.

### 2.1 Basic Description

Genetic algorithms are typically implemented as follows,
1. The problem to be addressed is defined and captured in an objective function that indicates the fitness of any potential solution.
2. A population of candidate solutions is initialized subject to certain constraints.

Typically, each trial is coded as a vector x, termed a *chromosome,* with elements being described as *genes* and varying values at specific positions called *alleles.* All solutions should be represented by binary strings. For example, if it is desired to find the scalar value x that maximizes
$$F(x) = -x^2,$$
Then a finite range of values for x would be selected and the minimum possible value in the range would be represented by the string [0... 0], with the maximum value being represented by the string [1... 1]. The desired degree of precision would indicate the appropriate length of the binary coding.

1. Each chromosome, $X_i$, i = 1,....,P, in the population is decoded into a form appropriate for evaluation and a fitness score is assigned to each chromosome, $\mu(x_i)$, according to the objective.
2. A probability of reproduction, $p_i$, i = 1,..,P, is assigned to each chromosome, so that its likelihood of being selected is proportional to its fitness relative to the other chromosomes in the population. If the fitness of each chromosome is a strictly positive number to be maximized, this is traditionally accomplished using **roulette wheel selection,** Figure 1.

3. According to the assigned probabilities of reproduction, p, *i=* I, *P,* a new population of chromosomes is generated by probabilistically selecting strings from the current population. The selected chromosomes generate offspring via the use of specific genetic operators, such as crossover and bit mutation. Crossover is applied to two chromosomes (parents) and creates two new chromosomes (offspring). This is done by selecting a random position along the coding and by exchanging the section that appears before the selected position in the first string with the section that appears after the selected position in the second string, and vice versa. Figure 2 shows this process. Bit mutation simply offers the chance to flip some bits in the coding of a new solution. Typical values for the probabilities of crossover and bit mutation range from 0.25 to *0.95* and 0.001 to 0.01, respectively.

4. The process is halted if a suitable solution is found or if the available computing time is expired. Otherwise, the process goes to step 3, where the new chromosomes are scored and the procedure iterates.

For example, suppose the task is to find a vector of 100 bits {0, 1} such that the sum of all the bits in the vector is maximized. The objective function could be written as

$$\mu(x) = \sum_{i=1}^{100} x_i,$$

Where x is a vector of 100 symbols from {0, 1}. Such vector x could be scored with respect to μ(x) and would receive fitness ranging from zero to 100. Let an initial population of 100 parents be selected completely at random and subjected to roulette wheel selection in light of μ(x), (fitness function) with the probabilities of crossover and bit mutation being 0.8 and 0.01, respectively. The process will rapidly converge to a vector of all 1's.

A number of issues must be addressed when using a genetic algorithm [1].
1. Using of binary representation versus floating point representation especially in real values optimization.
2. Selection in proportion to fitness can be problematic. Since (1) Roulette wheel selection depends on positive values, and (2) simply adding a large constant value to the objective function can eliminate selection. Other problems of fairness may arise for this selection technique.
3. Premature convergence is another important concern in genetic algorithms. This occurs when the population of chromosomes reaches a configuration such that crossover no longer produces offspring that can outperform their parents, as must be the case in a homogeneous population. Under such circumstances, all standard forms of crossover simply regenerate the current parents. Any further optimization relies solely on bit mutation and can be quite slow. Although many open questions remain, genetic algorithms have been used to address diverse practical optimization problems successfully.

**3 Genetic Algorithm for best training set generation and selection**

The motivation of this paper was to find a solution for the problem of Arabic character recognition. Neural network approach is considered as a good solution for the pattern recognition problem. Back-propagation networks form a very good structure that can actually solve our recognition problem. A training set that covers the problem space must be constructed. Many questions arose for the neural network approach, such as network structure, activation function, and initial weights. New question is to be asked here. What is the best training set to be used with the BPN (Back-Propagation Network)?
In the following an answer for this question is given.

## 3.1 Problem Formulation

The system is required to recognize typed characters of different styles and sizes. Size problem can be solved by normalization and zooming. The problem of different styles, fonts, is currently solved by training the network by all possible patterns or fonts. This is actually a good but not an optimal solution since it rises the training time due to the large training set volume. Another solution is to find a covering, yet small, set that represents all the candidate patterns. For the current problem, we are trying to find a training set that used to train a BPN, and then the network can recognize the whole target set. As described earlier, the parameter that governs the learning quality of the network is the residual error of the network. That is the root mean square error of the patterns when compared with the desired output, i.e. if the network recognizes all the patterns; the residual error for all patterns is under the acceptable error tolerance. Figure 6 shows some Arabic letters written in different fonts. The presented subset of characters is selected to represent all character shapes, as will be explained later, also the most popular writing fonts are considered.

Let us describe the network and give the solution without using optimization and GA (Genetic Algorithm). Each pattern is normalized and zoomed into a unified 16x16 window and digitized into a 256-bit vector. Our input layer size is going to be 256. The output layer size is 4 nodes since we have 12 different classes which can be decoded into 4-bit code. The hidden layer size is taken to be 32 nodes $=\sqrt{256*4}$ [2], this value is found in many neural network literature and is advised to be taken as an initial guess, and the network has one hidden layer. As mentioned earlier, the problem can be solved by training the network by all the patterns, but this solution is very time consuming. This solution was tried and a total time of about 2 hours was taken to complete the training process (the 108 character forms are considered as the training set). So our problem is to find the best training set to be used with a back-propagation network such that this trained network can recognize correctly any character in the above array.

## 3.2 Genetic Algorithm usage, a first proposal

Let us consider genetic algorithm to solve the previous problem. There are two main characteristics of genetic algorithm that led this development:
1. Genetic algorithm is an efficient search algorithm
2. Within the process of genetic algorithm, new solutions are found by combining the original solutions in such a way that these new solutions were not members of the original set of solutions, first generation.

Now we formulate the problem into genetic algorithm context.
A first representation can be:
1. Consider each training set to be a chromosome, i.e. each complete set of characters (each line in Figure 3) is considered to form a chromosome.
2. Each pattern within a chromosome is considered a gene
3. The genetic algorithm is then applied.

   The fitness function can be characterized by finding the residual error for all the given patterns after using the chromosome as a training set and use this value as the fitness score. It can be said that the network has learned to recognize the features describing the patterns of the given training set, chromosome. Then the network is used to recognize all the given original array of patterns. With each pattern the output of the network is obtained and compared to the desired output for this pattern and an error term is calculated. The maximum of those errors is used as the measurement of how the chromosome can describe the whole array of characters. The genetic algorithm through its crossover and mutation operations results a chromosome with minimum fitness. Iterations continue until a chromosome resulting in an error less than the tolerance value. Figure 4 below shows the representation of the problem.

## 3.3 Problems with the first proposal

From the previous discussion a suitable formulation of the problem to fit the genetic algorithm is addressed. With this approach (representing the pattern by a gene and the set by a chromosome) we remark,
1. Each pattern here is a gene, i.e. we are no longer dealing with binary GA.
2. Crossover operation will produce new chromosomes with different combination of the original characters, and this must assume that a combination of the original characters actually holds a solution, which is not always the case.
3. Mutation here must be done by inverting a gene i.e. getting the negative of an image, which is obviously has no role in producing a valid solution.

So we can see that another encoding scheme must be used to overcome these problems. In the next section an extension of the genetic algorithm is proposed to alleviate the mentioned problems.

## 3.4 Beings, An Extension to Genetic Algorithm

Deep insight into the problem, one can find that a pattern, or a character, has got a big role in the training problem. Considering a pattern as a gene and not participating in the generation of the training set eliminates an important part that can be very useful. Again borrowing the nature terms let us propose a new construct and call it a ***being.*** As natural beings, it is bigger than a chromosome construct. It is actually constructed of chromosomes, just like chromosomes are constructed of genes. The being is subjected to the following definition and conditions

1. Each Being has many chromosomes
2. Each chromosome has many genes
3. Each chromosome corresponds to a specific pattern and each gene corresponds to an image pixel
4. Pure beings have all pure chromosomes
5. Each being must have the same number and types of chromosomes (each being must have a pattern for each character)
6. A population of pure beings forms pure tribe.
7. Crossover of 2 beings happens on chromosomes of the same genetic type to produce
    - New beings with different combination of chromosomes (if the crossover point happens on a chromosome boundary)
    - New beings with hybrid chromosomes (if the crossover point happens inside a chromosome)
8. Mutation is carried out on genes and represents noise
9. Competition is between beings and not chromosomes.

Figure 5 shows the construction of a being.

Considering the above definition and conditions, then each character is encoded into a binary vector forming a chromosome. Each being is formed of twelve chromosomes where each chromosome corresponds to a character in the set. The problems of the first proposal are solved. Binary representation is used back again since each gene represent an image pixel and is either binary 0 or 1 for white or black pixels respectively. Crossing over operation may produce hybrid patterns constructed from original patterns and these new patterns hold the features of the original ones. In the same time we did not lose the pattern exchanging operation as with the first proposal. Mutation here when carried out on genes actually helps to represent noise which is always found when using optical scanners to gain the characters images. By this proposal the process can be carried out to find the best training set by the same fitness function used before, namely, minimum residual error after training. The training for each being is started from the same set of initial weights. And the training continues until all the chromosomes, characters, of the being are recognized. Then the original set is tested for fitness calculation.

Figure 6 shows the in-chromosome crossing over operation and shows how new patterns are generated from the original patterns.

Note that selection actually happens on normalized patterns of the same size.

Figure 7 shows the result of such crossing over.

**3.5 Application of the proposed algorithm to the character recognition problem**

The proposed algorithm was tested with the given set in Figure 3 with the aim to find a set that has one pattern for each character in concern and that must contains the features that describe all the given characters as described earlier. The neural network used is as follows

Network type:                         Back-Propagation
Number of hidden layers:              1
Input layer size:                     256
Hidden layer size:                    32
Output layer size:                    4
Activation function:                  Sigmoid function in its basic shape
Tolerance                             0.05

For each being used, the network is initialized with the same set of weights and training continues until all being chromosomes are recognized.

The genetic algorithm parameters are

Being size:                           12 chromosomes
Chromosome size:                      256 gene
Maximum number of generations:        40 generations
Crossover probability:                0.25
Mutation probability:                 0.1
Population size:                      9 beings

Single point crossover is used and the initial population, generation one, was taken to be the original training set, all pure beings.

The system is simulated and Figure 8 shows the residual error in each generation for the best being.

As can be seen, the best result was obtained in generation 12 and was about 0.02 (residual error). This value is actually less than the tolerance, which was 0.05, but the experiment allowed the process to continue until the $40^{th}$ generation to test the performance.

Figure 9 shows the resulting which realizes the minimum fitness

Comparing the resulting set to the original array, we find that actually each character holds the important features of all the corresponding characters in the original array.

The algorithm actually works and takes about 40 minutes to complete the process.

We cannot take the time gain as actually a real gain, since the genetic algorithm is probabilistic by nature. If we were going to stop the process when a good satisfactory solution is achieved, much less time, in our case, is realized. The actual benefit of such a process is that we can improve the convergence of the neural network in the recognition process to a valid solution when a small sized training set is used. For large training sets, that assumption may not always hold. What we actually done is to help the neural network in selecting the best features to describe the problem space. In the same time we cannot rely completely on the automatic feature selection nature of a neural network to do the job.

### 3.6. Defending the proposed extension

Now let us say, Why bother with a new extension and definitions, Why not just view the problem as taking a large size chromosome where parts of this chromosome represents patterns and apply the normal genetic algorithm. Actually this point of view arose while developing the work.

While the extension was just thoughts, one operation of the genetic algorithm was concerned very much. That was crossover operation. By our proposal two different crossing over operation can be defined and they are mentioned above,

In-chromosome crossover and boundary-crossover. Variation of these operations provides situation suitable for different applications. Although single point crossover was used in the problem at hand which result in either of the two mentioned crossover operation, redefining the algorithm to have both operations in the same time can affect the performance of the system an it is a good point of further studies. The proposed extension is clearer and there are direct correspondences between problem blocks and algorithm constructs.

### 4. Conclusion

This paper introduced a novel way for constructing the best training set for a back-propagation network used for Arabic character recognition.

A new extension to the basic genetic algorithms is also introduced to scale the algorithms up to fit the problem of character set selection.

### 5.References.

**6.Figures.**

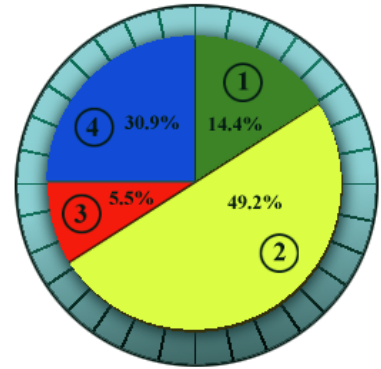

| Number | String | Function | % total |
|--------|--------|----------|---------|
| 1 | 011110110111 | 169 | 14.4 |
| 2 | 011010010111 | 576 | 49.2 |
| 3 | 101101111011 | 64 | 5.5 |
| 4 | 111010010111 | 361 | 30.9 |
| Total |  | 1170 | 100 |

Figure 1. Roulette wheel selection mech

```
                Crossover Point
Parent #1:      1101 | 0111101      Offspring #1: 10100111101
Parent #2:      1010 | 0000100      Offspring #2: 11010000100
```

Figure 2. The one-point crossover operator applied to two parents.

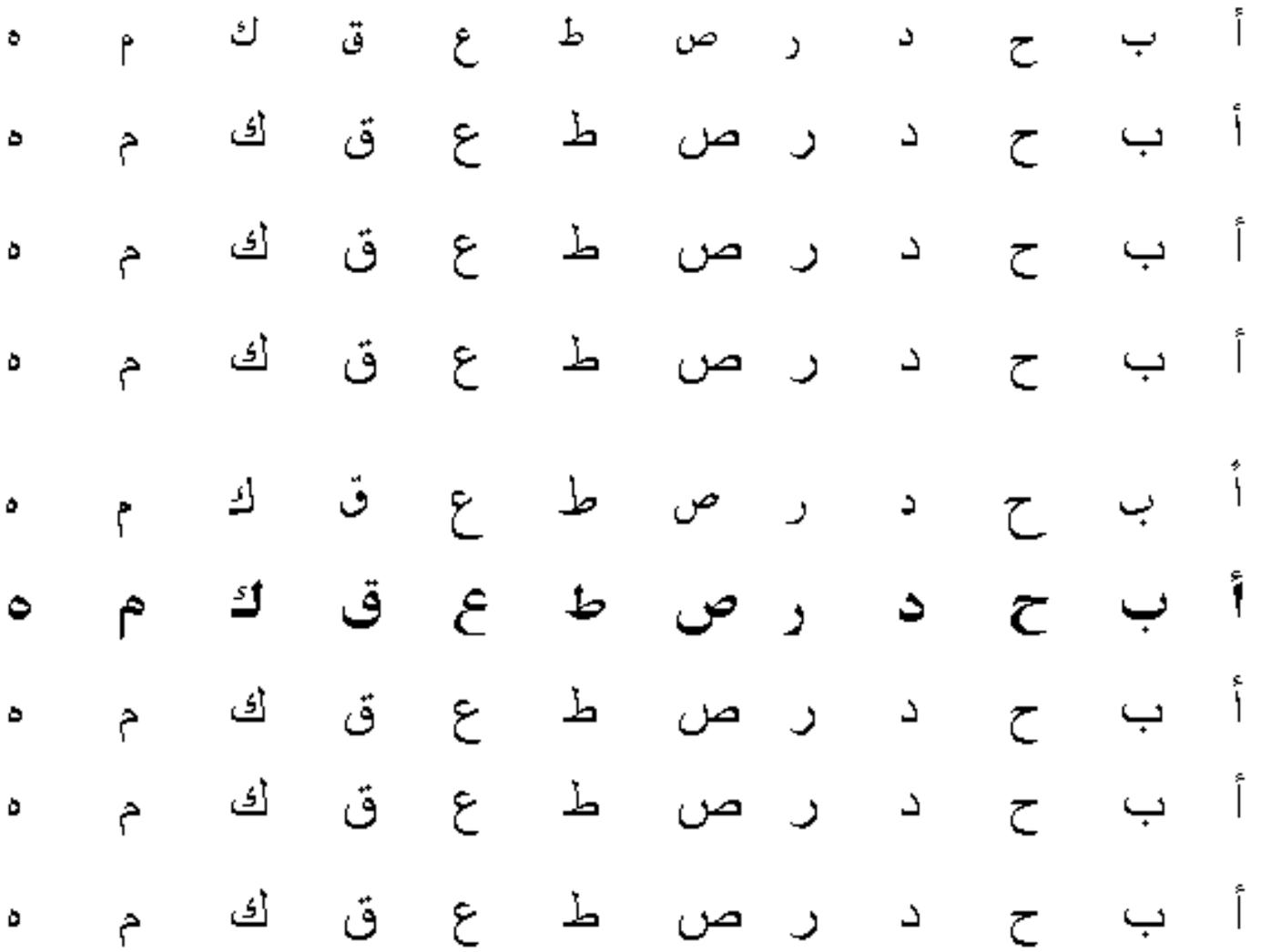

Figure 3. The considered subset of Arabic characters written in different fonts.

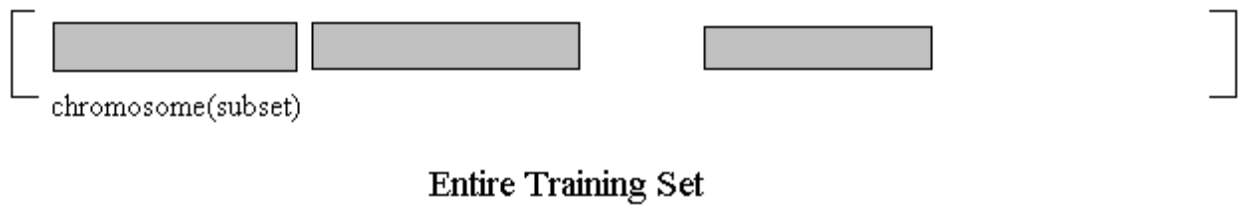

Figure 4. Entire training set divided into chromosomes

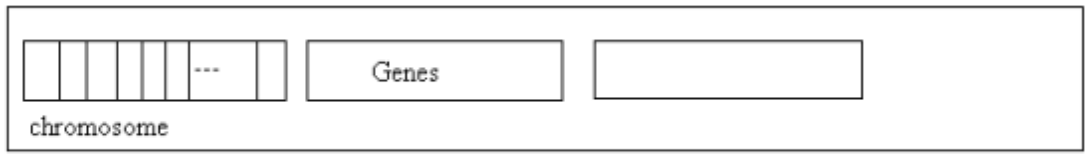

Figure 5. A Being

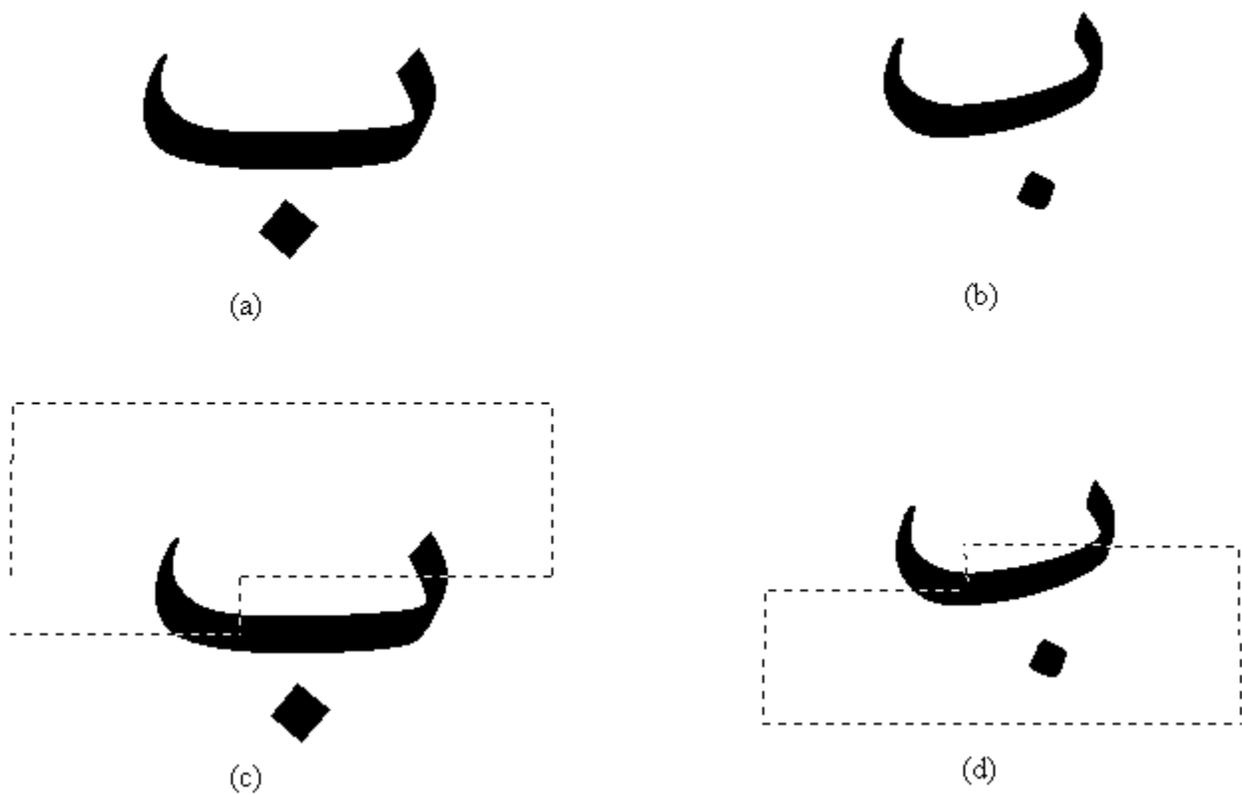

Figure 6 (a) and (b) original patterns of the same character "Ba" in Arabic (c) and (d) the selected portions of each image for crossing over

ب

Figure 7. Result of in-chromosome crossover operation.

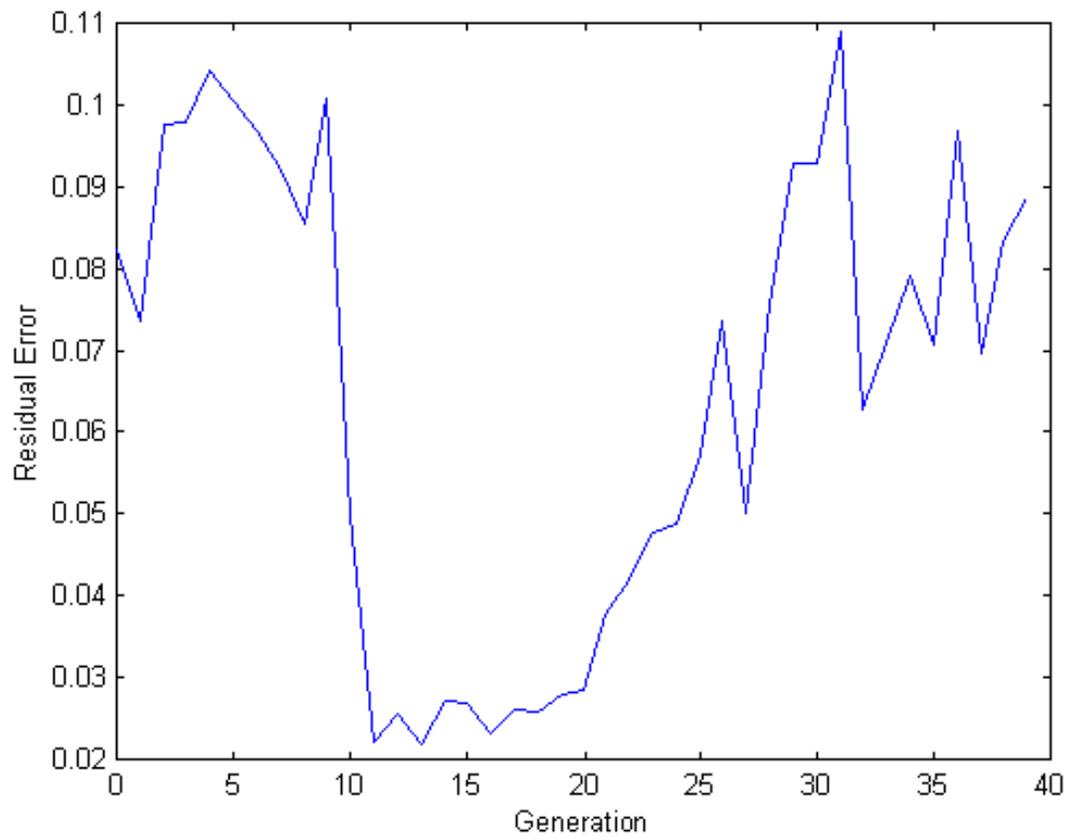

Figure 8. Progress of fitness of beings.

ا ب ح د ر س ص ط ع ق ك م ه

Figure 9 Best training set